\documentclass[conference]{IEEEtran}
\IEEEoverridecommandlockouts
\usepackage{cite}
\usepackage{amsmath,amssymb,amsfonts}
\usepackage{algorithmic}
\usepackage{graphicx}
\usepackage{textcomp}
\usepackage{xcolor}
\usepackage{algorithm}
\usepackage{amsmath}
\def\BibTeX{{\rm B\kern-.05em{\sc i\kern-.025em b}\kern-.08em
    T\kern-.1667em\lower.7ex\hbox{E}\kern-.125emX}}
\begin{document}

\title{Causal Explanations from the Geometric Properties of ReLU Neural Networks\\
}

\author{\IEEEauthorblockN{1\textsuperscript{st} Hector Woods}
\IEEEauthorblockA{\textit{Department of Computer Science} \\
\textit{University of York}\\
York, United Kingdom \\
hjvw500@york.ac.uk}
\and
\IEEEauthorblockN{2\textsuperscript{nd} Philippa Ryan}
\IEEEauthorblockA{\textit{Department of Computer Science} \\
\textit{University of York}\\
York, United Kingdom \\
philippa.ryan@york.ac.uk}
\and
\IEEEauthorblockN{3\textsuperscript{rd} Rob Alexander}
\IEEEauthorblockA{\textit{Department of Computer Science} \\
\textit{University of York}\\
York, United Kingdom \\
rob.alexander@york.ac.uk}
}

\maketitle

\begin{abstract}
Neural networks have proved an effective means of learning control policies for
autonomous systems, but
these learned policies are difficult to understand due to the black-box nature of
neural networks. This lack of interpretability makes safety assurance for such
autonomous systems challenging.
The fields of eXplainable Artificial Intelligence (XAI) and eXplainable
Reinforcement Learning (XRL) aim to interpret the decision-making processes of
neural networks and autonomous agents, respectively. In particular, work on
causal explanations aims to provide ``why" and ``why not" explanations for why a
model made a given decision. However, most of the work on explainability to date
utilises a distilled version of the original model. While this distilled policy is
interpretable, it necessarily degrades in performance significantly when compared to the
original model, and is not guaranteed to be an accurate reflection of the
decision-making processes in the original model and as such cannot be used to
guarantee its safety.
Recent work on understanding the geometry of ReLU neural networks shows
that a ReLU network corresponds to a piecewise linear function divided into
regions defined by an n-dimensional convex polytope. Through this lens, a
neural network can be understood as dividing the input space into distinct
regions which apply a single linear function for each output neuron.
We show that this geometric representation can be used to generate causal
explanations for the network's behaviour similar to previous work, but which
extracts rules directly from the geometry of Neural Networks with the ReLU
activation function, and is therefore an accurate reflection of the network's
behaviour. An implementation of our algorithm is available at https://github.com/HJWoods/ReLUExplanations
\end{abstract}


\section{Background}

\subsection{Introduction to Explainability for Deep Learning and
Reinforcement Learning}
eXplainable Artificial Intelligence (XAI) aims to provide insight into the
decision-making processes of AI systems, such as neural networks. In the context
of a control policy learnt via Reinforcement Learning (RL), the goal of
eXplainable Reinforcement Learning (XRL) is to verify that the system will not
enter an unsafe state regardless of the input and to promote trust in the system
by allowing users to gain meaningful insights about what decisions the model
will make for a given input and why it makes those decisions. This field is
distinct from, but related to, formal verification of AI/RL systems: the former
aims to provide guarantees that the system will not produce an unsafe output
given certain constraints, while the latter aims to uncover the decision-making processes of AI systems and present them to the user succinctly. Explainability remains useful even for a system that is guaranteed to be safe; as can be argued without the ability to ask the system why it made certain decisions, users will not necessarily trust the system based on guarantees from an external regulator alone \cite{xrl_survey}.

\subsection{What makes a good explanation?}
Given the early stage of explainability research, what exactly generating a "good" explanation entails is not completely understood, and there is a clear need to verify these explanations against real users.
Nevertheless in this section we will discuss some of the popular paradigms in explainability research with regard to tailoring explanations to the needs of end users.
\subsubsection{Model Reconciliation}
\cite{reconcile} consider explainability as a problem of model reconciliation; under this model the system
and the user both have distinct models of the environment in the system in which the system operates and an understanding of how the
system should behave in a given scenario, known as a "world model". These models do not necessarily align, and in fact
in the vast majority of cases we should expect that they will not align \cite{aligningrepr}. The
goal of explainability is to reconcile the user's model of how the system \textit{should}
behave with how it actually behaves. This allows the user to verify that the
system will produce a safe output for a given set of inputs, and if this is not the
case, to flag the issue to the designer of the system to resolve the issue.
Explainability also enables the user to identify cases where the model made an
action which is safe but nonetheless not the one expected by the user, in which case the user is able to understand why the decision made by the system was safe even though it defied their expectations.
\subsubsection{Human Factors Considerations}
Applying ideas from human factors research, [1] suggest that explainable systems should aim to maximize the
``likeliness" and ``loveliness" of a given explanation. Likeliness refers to the
probability that the given explanation adequately explains an input X, expressed in terms of probability: An explanation $E_i$
that maximizes the posterior probability $P(X|E_i)$ for an event X is
said to be an explanation with good likeliness. Loveliness refers to how
``satisfying" the explanation is to an end user. An agent could give an
explanation for an event with thousands of steps that gives a comprehensive
explanation for its thought process, but this explanation is of little use to the
user as they are unable to comprehend this much information or in terms of only its own parameters, which would be inaccessible to someone without a mathematical background. One of the most
important factors in loveliness then is simplicity - how easily the explanation can
be understood by an end user without intimate knowledge of AI/ML concepts and often even
of the domain in which the system is applied.
While loveliness can refer to a diverse range of approaches by which an
explanation can be made more satisfying, one of the most important aspects is
its conciseness; an explanation that is too long is inherently unsatisfying and, if too long, may be completely incomprehensible to a human reader. A good
explanation should therefore be as comprehensive as possible while being
composed of the minimum amount of information necessary for comprehension.
This is known as the Minimally Complete Explanation (MCE). While there are
other means by which explanations can be made more satisfying, particularly tailoring explanations to those without domain-specific knowledge, finding the
MCE is a good heuristic for generating satisfying explanations given that it
provides exactly enough information to explain why the system made a given
decision without overloading the user with extraneous information not relevant
to that decision.

\subsubsection{Causal models and counterfactual explanations}

Human factors research has shown that human beings tend to construct causal
models of the world: where each outcome is understood as being caused by some
other events/input variables \cite{causal}.
According to this model, we can represent a system's behaviour in terms of
its input variables, and in particular what conditions over the input variables
encouraged that decision, and why that decision was chosen over
any alternative.
This is represented by factual and counterfactual explanations; the former
provides an explanation in terms of the input variables as to why the model
made a given decision, and the latter explains why the model made a given
decision and not a counterfactual decision given the same input. In everyday
speech, we might express this as ``Why" or ``Why not" questions on the model's
behaviour \cite{causal}.
Previous work has utilized these factual/counterfactual explanations to allow
users to query the model as to why it made a certain decision for an
input/output pair - a ``why" explanation, and query for a given input why the
model did not produce a counterfactual output - a ``why not" explanation.
Explanations of this form allow the user to engage in ``dialogue" with the
system; through querying the system with different input/output pairs and
retrieving factual/counterfactual explanations, the user gains insight into the
model's behaviour, aligning their understanding of how the model works and its
actual behaviour.
However, applying this approach to a deep-learning based system is
challenging due to the black-box nature of neural networks. To date this
approach has been limited to simpler Reinforcement Learning algorithms such as
Q-learning \cite{causal}, hand-coded algorithms \cite{causalnavigation}, decision trees or other naturally interpretable models.
While previous work demonstrates the feasibility of this approach for
distilled versions of deep neural networks, we will demonstrate how to generate
explanations for neural networks consisting of only linear layers and the ReLU
activation function, applying insights from the geometric properties of such
networks.

\subsection{ReLU networks and polytopal decomposition}

The geometric properties of ReLU networks have attracted interest from a formal verification perspective and, in particular, the desire to find and enumerate the reachable
regions of neural networks. By obtaining the full set of reachable outputs and
the input regions they correspond to, it is possible to perform automatic safety
verification of the system by confirming that all input/output pairs satisfy
certain conditions \cite{facetvertex}\cite{reluplex}\cite{deeppoly}. In the case of ReLU neural networks, we can achieve this by exploiting the \textit{piecewise linear} nature of the function; within each linear region each layer is highly interpretable, given that it corresponds to a single linear transformation on the input space \cite{polydecomp}.
Consider a neural network consisting of only linear layers and the ReLU activation function between each layer of arbitrary depth $l$.
The ReLU activation is a piecewise linear function defined as:
\begin{equation*}
\max(x, 0)
\end{equation*}
The output of the $i$th linear/relu pair in a layer with the weights matrix $W$ and bias vector $b$ is therefore given by 
\begin{equation*}
\max(W_i x + b_i, 0)
\end{equation*}
Where $W$ is an $n \times m$ matrix holding the weights for that layer and $b$ is the bias
vector with $m$ elements. We can define the max function as 
\begin{equation*}
\max(x, y) = 
\begin{cases}
x & \text{if } x > y \\
y & \text{otherwise}
\end{cases}
\end{equation*}
The output for each neuron in layer $i$ of a network $N$ is therefore given by: 
\begin{equation}
\text{ReLU}(N_{i}(x)) = 
\begin{cases}
w_i x + b_i & \text{if } w_i x + b_i > 0 \\
0 & \text{otherwise}
\end{cases}
\end{equation}
It is possible to express the output of each neuron as a single linear
inequality and therefore each neuron divides the input space by a hyperplane,
creating two half-spaces for each neuron; one where it outputs $0$ and one where it
performs the linear transformation $w_i x + b_i$ over the input. Given that the
output as a whole is simply a vector which contains the outputs of each
individual neuron, we can represent each layer as a system of linear inequalities that describe the activation patterns of each neuron following the ReLU function. Each unique combination of zero/non-zero activations is described by a unique system of linear inequalities $H$ in the standard form $Ax <= b$, where
each row of A and b is of the form:
$$-w_{i}x <= b_{i}$$
when the activation $w_ix$ + $b_i$ is greater than 0, and
$$w_{i}x <= -b_{i}$$
when the activation $w_ix$ + $b_i$ is less than or equal to 0, and hence exactly 0 following the ReLU function.

The standard form of a system of linear inequalities corresponds exactly to the H-representation of a convex polytope in
$D$-dimensional space. This polytope $P$ describes the exact region in the input space where the activations of layer $i$ will correspond to $H$. Each possible combination of zero/non-zero activations, and therefore each polytope $P$ for layer $i$ is therefore represented by a
unique system of linear inequalities. Intuitively, the upper bound on the number
of possible output polytopes for a layer with $n$ neurons is $2^n$, though in practice many of these inequalities are infeasible.  Given this fact, as suggested in \cite{polydecomp} we can represent each polytope as a collection of bit vectors for each layer, where the bit vector of an input $x$ for the layer $i$ with $n$ neurons is given by:
$$s_i(x) = [bit_{w_{i1},b_{i1}}(x), bit_{w_{i2},b_{i2}}(x), bit_{w_{in},b_{in}}(x)] $$
where
\begin{equation*}
bit_{w_{i,j},b_{i,j}}(x)
\begin{cases}
1 & \text{if } w_{ij}x + b_{ij} > 0\\
0 & \text{otherwise}
\end{cases}
\end{equation*}
\cite{polydecomp}

We can further stack each bit vector $s_i$ to obtain a single bit vector $s$ which represents the zero/non-zero activations of the network as a whole. Each bit vector $s$ represents a single polytopal region in terms of the input space that produces the activation pattern given by $s$. Within the confines of a
polytope, each neuron in the output layer is given by a single linear
transformation on the input $x$. The network is therefore highly interpretable within the bounds of each polytope, as the network is reduced to a single linear transformation for each output neuron. We can obtain the output of each neuron with respect to an activation pattern by expanding the bit vector $sL$ for a layer $L$ such that when multiplied by $w$ the vector is applied as a mask to the output of each neuron. The output for an input x is given by:
\\ \\
$G(x) = w_{L+1}diag(s_L)\hat{w}_{L}(x)+w_{L+1}diag(s_L)\hat{b}_{L}+b_{L+1}$
\\
where $diag(s_L)$ is a diagonal matrix of $s_L$.
\cite{polydecomp}
\\

In the context of a control policy learnt by a neural network, such as a Deep-Q network, each decision the network can make at a time point $t$ is typically given by the highest value of the output class following some activation function $\sigma$
where $\sigma \neq ReLU$. $\sigma$ is most commonly the $softmax()$ function. The decision made by the network for an input x within the confines of a polytope $P$ is therefore given by $argmax(\sigma(G(x))$.

\cite{facetvertex} \cite{skelex}\cite{vertexreachability} demonstrate how to utilise this geometric property of ReLU networks to identify all of the reachable outputs for all possible inputs to the neural network. This can then be used to verify the network's safety by ensuring that all input $\Rightarrow$ output mappings do not violate any constraints that imply the action the system took was unsafe.

The main issue with this approach is the exponential complexity associated with enumerating all of the output polytopes. Given the bit vector representation, where each neuron is represented by a either 0 or 1, there are $2^{n}$ possible bit vectors and this forms the theoretical bound for the number of polytopes that describe the network's behaviour. While \cite{facetvertex} implement a parallelised algorithm that can feasibly find all of the polytopes for networks with thousands of neurons, most modern neural networks will have an order of magnitude more neurons than this.

\subsection{Adjacent Polytope Marching}

Given the bit-vector representation of each polytope as described in \cite{polydecomp}, adjacent polytopes can be identified by flipping any of the bits in the bit vector. This identifies a polytope which lies on the other side of a linear inequality. This means that the distance between polytopes in terms of satisfied linear inequalities is simply the Hamming distance of their bit vectors, and adjacent bit vectors have a Hamming distance of 1. \cite{reachablemarching} \cite{traversingpolytopes} demonstrate the feasibility of utilising this property to explore adjacent regions to a given input. This can be used to identify \textit{connected} regions of the input space and identify partial geometric representations of the network. While the theoretical number of polytopes for a ReLU network is $2^n$, many of these polytopes will be exceedingly rare or only activate for input regions which are theoretically possible but far outside the real operating range of the system. Polytope marching can therefore be used to find the practically reachable regions of a neural network and avoid the exponential complexity associated with enumerating all possible polytopes.

\subsection{Polytope Decomposition for Explainability}
Villani et. al \cite{pice} present Polyhedral Complex Informed Counterfactual Explanations (PICE): a method for generating provably minimally complete counterfactual explanations for ReLU networks, based on an approach which generates the full polytopal decomposition for a neural network. By obtaining the full decomposition, they are able to generate explanations which are \textit{exact}, i.e. a 100\% accurate representation of the network's behaviour on any given input. The main limitation with this approach is the process of decomposing the network itself; as previously discussed, for a network with \textit{n} neurons can produce an upper bound of $2^n$ unique regions, in the case that all hyperplanes created by the neurons intersect. While numerous works have observed that the number of feasible regions is much lower in practice \cite{surprisinglyfew}\cite{polydecomp}, (a result which is consistent with expectations from hyperplane arrangement theory \cite{hyperplanearrange}) the number of regions for even a very small network remains very high, and the process of uncovering them is computationally expensive. The best known method for enumerating all polytopes of a ReLU neural network to date is given by Balestriero \& LeCun \cite{fastandexact} who demonstrate a method for generating a full decomposition which has linear time complexity with respect to the number of feasible linear regions, but requires solving many linear programs. As such, generating full decompositions for even small networks is generally speaking infeasible. Given this limitation, Villani et. al also provide a method for generating approximate explanations based on the PICE approach, \textit{PICE-Fast}, which is non-exact but able to provide a lower bound on the probability of the explanation being accurate.

\section{Method}

In this section we will demonstrate how the intuition behind polytopal
decomposition of ReLU networks can be used to generate why/why not
explanations for a neural-network based control policy learnt via Reinforcement Learning. In a similar vein to PICE, our method is also capable of generating exact explanations, but by contrast rather than aiming to find the full decomposition we instead utilize a polyhedral marching approach, starting at the polytope containing a point x until we encounter a polytope containing the counterfactual class, i.e. enumerating only the polytopes which are immediately relevant to the explanation.
\subsection{``Why" explanations}
\begin{algorithm}
\caption{Generate a ``Why" Explanation for an input x}
\begin{algorithmic}[1]
\STATE \textbf{Input}: Neural network $N$ with $l$ layers, input $x \in \mathbb{R}^n$, where the output layer uses the activation $\sigma$ and all other layers use ReLU
\STATE \textbf{Output}: Minimally complete explanation for output $N(x)$
\STATE Forward pass: $x \rightarrow N$ and obtain output $N(x)$
\FOR{each layer $i$ from $1$ to $l$}
    \STATE Collect activation vector $A_i$
    \STATE Create binary pattern $sL_i$ where $sL_{ij} = 1$ if neuron $j$ is active, $0$ otherwise
\ENDFOR
\STATE Combine all patterns to form activation signature $s = [sL_1, sL_2, \ldots, sL_l]$
\STATE Construct H-representation of polytope $P$ corresponding to signature $s$ as system of linear inequalities
\STATE Identify output class $c = argmax(\sigma(N_l(x)))$
\STATE Form subset polytope $P_{\text{Output}} \subset P$ that produces output class $c$
\STATE Formulate linear program to eliminate redundant constraints using the approach in \cite{polydecomp}:
\STATE \hspace{1em}For each constraint $i$ in the system $Ax \leq c$, solve:
\STATE \hspace{2em}maximize $a_i^T x$
\STATE \hspace{2em}subject to: $\tilde{A}x \leq \tilde{c}$ (where $\tilde{A}$ and $\tilde{c}$ exclude row $i$)
\STATE \hspace{1em}If the optimal value $\leq c_i$, constraint $i$ is redundant and can be removed
\STATE Extract simplified constraint set $H'$ after removing all redundant constraints
\STATE \textbf{return} $H'$ as the minimally complete explanation
\end{algorithmic}
\end{algorithm}
Consider an input $x \in \mathbb{R}^n$ to a neural network $N$ consisting of $l$ linear
layers followed by the ReLU activation function.
We pass the input $x$ to $N$ and collect the activations for each layer $i$. This
will be a sparse vector $A_i$ where each element $A_{ij}$ is either a positive real number
or $0$. From this we can construct the bit vector $sL_i$. We can stack each $sL_i$ to
obtain a single vector $s$ which describes the activations of $N$ for the input $x$.

As explained in \cite{polydecomp}, each neuron in the final layer of the network (and therefore each output class) corresponds
to a single linear transformation given by $G(x)$. The
input $x$ is a subset of the convex polytopal region $P$, where $P$ is given by the
system of linear inequalities $H$.
For a classification problem, the output class is given by $c = \arg\max(\sigma(N_l(x)))$ where $\sigma$ is an arbitrary activation function.

To explain the output class within this polytopal region, we define a subset polytope $P_{\text{Output}} \subset P$ that represents all points within $P$ where class $c$ has a higher activation than all other classes. We do this by appending the following inequalities to the system defining $P$:

$$\sigma(N_{lc}(x)) > \sigma(N_{lj}(x)) \text{ for all } j \neq c$$

This derives a second polytope $P_{\text{Output}}$ which is a subset of $P$ where the chosen class $c$ has a higher activation than all other classes.

The linear system that describes $P_{\text{Output}}$ may contain redundant constraints
(and is likely to do so in lower input dimensions). Following the approach of Liu et. al \cite{polydecomp}, we can remove these redundant constraints by solving a series of linear programs. 

Let the system of linear inequalities be represented as $Ax \leq c$, where $A = [a_1, a_2, \ldots, a_h]^T$ and $c = [c_1, c_2, \ldots, c_h]^T$ with $a_i \in \mathbb{R}^n$ and $c_i \in \mathbb{R}$. To determine if the $i$-th constraint $a_i^T x \leq c_i$ is redundant, we first define:

$$\tilde{A} = [a_1, a_2, \ldots, a_{i-1}, a_{i+1}, \ldots, a_h]^T$$
$$\tilde{c} = [c_1, c_2, \ldots, c_{i-1}, c_{i+1}, \ldots, c_h]^T$$

Then we solve the linear program:

$$\text{maximize} \quad a_i^T x$$
$$\text{subject to} \quad \tilde{A}x \leq \tilde{c}$$

If the optimal objective value is less than or equal to $c_i$, then the $i$-th constraint is redundant and can be removed. This is because the remaining constraints already ensure that $a_i^T x \leq c_i$ will be satisfied.

We determine the minimal set of constraints $H'$ by iteratively applying this process to remove all redundant constraints. The simplified system of linear inequalities $H'$ can be considered a minimally
complete explanation for $N(x)$ in that it contains only the constraints which are
relevant to defining the regions $P$ and $P_{\text{Output}}$.

\subsection{Why not explanations}

\begin{algorithm}
\caption{Generate a ``Why Not" Explanation for an input x and a counterfactual output O'}
\begin{algorithmic}[1]
\STATE \textbf{Input}: Neural network $N$ with $l$ layers, input $x \in \mathbb{R}^n$, counterfactual output class $c'$, where the output layer uses the activation $\sigma$ and all other layers use ReLU
\STATE \textbf{Output}: Minimally complete explanation for why $N(x) \neq c'$
\STATE Identify actual output class $c = \text{argmax}(\sigma(N_l(x)))$
\STATE Follow steps 3-11 from Algorithm 1 to obtain activation signature $s$, polytope $P$ and subset polytope $P_{\text{Factual}} \subset P$
\STATE Form subset polytope $P_{\text{Counterfactual}} \subset P$ with additional constraints:
\STATE \hspace{1em}$\sigma(N_{lc'}(x)) > \sigma(N_{lj}(x))$ for all $j \neq c'$
\STATE Check feasibility of $P_{\text{Counterfactual}}$ by solving the linear program:
\STATE \hspace{1em}Maximize x subject to $Ax \leq b$, where $A$ and $b$ define $P_{\text{Counterfactual}}$
\IF{$P_{\text{Counterfactual}}$ is feasible}
    \STATE Extract weight vectors $w_c$ and $w_{c'}$ for output classes $c$ and $c'$
    \STATE \textbf{return} ``Output $c$ was chosen over $c'$ because $(w_c - w_{c'})^T\text{diag}(s)x + (b_c - b_{c'}) > 0$"
\ELSE
    \STATE Initialize distance $d = 1$
    \WHILE{true}
        \STATE Find all bit vectors $s'$ with Hamming distance $d$ from $s$
        \FOR{each bit vector $s'$}
            \STATE Construct polytope $P'$ corresponding to $s'$
            \STATE Form subset polytope $P'_{\text{Counterfactual}} \subset P'$ with constraints for class $c'$
            \STATE Check feasibility of $P'_{\text{Counterfactual}}$
            \IF{$P'_{\text{Counterfactual}}$ is feasible}
                \STATE Identify the positions where $s$ and $s'$ differ
                \STATE Extract the corresponding constraints from $P$ and $P'$
                \STATE \textbf{return} These constraints as the explanation for ``Why not $c'$"
            \ENDIF
        \ENDFOR
        \STATE $d \gets d + 1$
    \ENDWHILE
\ENDIF
\end{algorithmic}
\end{algorithm}

Consider a known input-output pair $x$, $N(x)$ and a counterfactual output $N(x)'$.
The goal in explaining why $N(x)$ and not $N(x)'$ is to find the minimum set of
constraints that differentiate the two output classes in the context of the input
$x$.
As in the ``Why" explanation, we first pass $x$ to the model, collect its
activations and construct the bit vector and system of linear inequalities for the
polytope $P$.

We then find two subset polytopes of $P$: $P_{\text{Factual}}$ and $P_{\text{Counterfactual}}$ that
represent the constraints for the real and counterfactual output respectively. $P_{\text{Factual}}$ 
is defined by all points in $P$ where:

$$\sigma(N_{lc}(x)) > \sigma(N_{lj}(x)) \text{ for all } j \neq c$$

And $P_{\text{Counterfactual}}$ is defined by all points in $P$ where:

$$\sigma(N_{lc'}(x)) > \sigma(N_{lj}(x)) \text{ for all } j \neq c'$$

Where $c$ is the actual output class and $c'$ is the counterfactual class.

In some cases, $P_{\text{Counterfactual}}$ may not be a feasible region; i.e. there are no points that satisfy the system of linear inequalities for $P_{\text{Counterfactual}}$ and the output N(x)' is not possible for inputs in this polytope.
To verify whether this is the case, we solve the linear program:
$$\text{maximize }  x$$
$$\text{subject to} \quad Ax \leq c$$
where $A$,$c$ are the constraints that define the polytope $P_{\text{Counterfactual}}$.

If $P_{\text{Counterfactual}}$ is feasible, then the explanation is simply that the output class $N(x) > N(x)'$, or in terms of the weights of the model:
$$\sigma(G_i(x)) > \sigma(G_{i'}(x))$$

If this is not the case, then our goal is to find the nearest polytope to $P$ that
has feasible inputs for the output class $N(x)'$. We can achieve this by utilizing
the Adjacent Polytope Marching algorithm described in \cite{reachablemarching}. Starting with the bit
vector that describes $P$, we flip the bit vector to find a polytope $P_{\text{Adjacent}}$
which satisfies all but one of the same inequalities as $P$. $P$ is adjacent to
$P_{\text{Adjacent}}$ in that they have a distance of one and lie on alternate sides of
exactly one of the hyperplanes of $H$.
For the adjacent polytope we once again generate the subset $P_{\text{Counterfactual}}$
and verify whether it is feasible. We continue this process, enumerating all
adjacent polytopes to $P$, increasing the allowable Hamming distance if the
directly adjacent polytopes don't contain $N(x)'$. Once we find the polytope $P'$
which contains $N(x)'$, we can present the explanation ``Why not $N(x)'$" as the
constraints which distinguish $P$ from $P'$. This is given by the difference in the Hamming distance of the two polytopes.
An explanation of this form is minimally complete in that it consists of the
minimum set of constraints that distinguish $N(x)$ from $N(x)'$.

\subsection{Polytope Explanations in V-Representation}

Given the dual representations of convex polytopes, we can provide explanations
as either the series of half-space intersections that define each polytope
(H-Representation) or the polytope's extreme points (V-Representation). The
V-Representation provides the minimum/maximum values for the input in each
dimension, or in other words the range of inputs that satisfy the polytope. This representation offers greater brevity than the H-Representation, which scales in size both with respect to the dimensionality of the input space and the size of the network.

In the context of a ``Why" explanation, we can find the extreme points of the
input's $P$, and the subset of that polytope $P_{\text{Output}}$) which contains all inputs that result in the output class $N(x)$. This provides two input ranges; one which defines the local region around the point $x$, and one which defines the range of inputs within that region which are classified as $N(x)$. In the context of ``Why not" explanations, we are given two local regions, $P$, which contains x, and $P'$ which is the closest region to $P$ that can produce the output class $N(x)'$, where P may equal P'.

\section{Discussion and Limitations}

\subsection{H-Representation versus V-Representation for Polytope
Explanations}

The V-Representation is arguably more interpretable given that it provides the
lower and upper bounds for each polytope and the bounds in which it will
make a given decision for inputs within that polytope. However, it doesn't
explain how the model arrived at this polytope, or in other words the decision
making processes of earlier layers, while the H-Representation is superior in this respect. Determining the V-representation also introduces exponential space/time complexity to the algorithm in that a D-dimensional hypercube has $2^D$ vertices, and this forms the upper bound for the number of vertices in a D-dimensional convex polytope. Nevertheless determining the V-Representation for low
dimensional input spaces is entirely feasible and can be used to supplement the
H-representation explanation. Control policies tend to operate on few
dimensions, while data-driven AI such as images or text has many tens or
hundreds of dimensions.
It could be argued that the H-Representation exhibits poor scaling with
respect to the size of the model, as each neuron adds an additional constraint to
the definition of each polytope, and as such the length of explanations scales
linearly with the size of the model. The V-Representation has a constant size and
doesn't scale with the size of the network, but exhibits exponential scaling with
respect to the number of input dimensions.
The V-Representation, when it is feasible, offers a much clearer and more
concise explanation for the network's behaviour, but is only feasible for very
small input dimensions.

\subsection{Time/Space Complexity for generating explanations}
The computational complexity using our method varies between ``Why" and ``Why not" explanations.

For ``Why" explanations, the output polytope can be found in $O(1)$ time, given that this is equivalent to simply passing the input through the network. To simplify the output polytope, the algorithm then needs to solve n linear programs to identify the redundant constraints. The time complexity for generating a ``Why" explanation for a given input is therefore $O(n)$ in both the best and worst case.

``Why not" explanations are also generated in $O(n)$ time in the best case, in the scenario where the first polytope contains a feasible region for the counterfactual output. However, in the worst case the time complexity for the adjacent polytope marching algorithm is $O(2^n)$, where n is the number of neurons in the network. In the case of worst-case time complexity, the task of generating the explanation is equivalent to finding the full polytopal decomposition of the original network. However in practice, this upper bound is extremely unlikely, and in practice the algorithm is likely to terminate quickly, given that only a single polytope need contain a feasible point for the algorithm to halt.

Given that each linear program is independent, both algorithms are highly parallelizable, and this somewhat overcomes the exponential complexity associated with generating ``why not" explanations.

Both algorithms have constant space complexity, given that ``Why" explanations store only the output polytope, and ``Why not" explanations store only the factual polytope, and the current counterfactual bit vector/polytope.

\subsection{H-Representation and Sparsity}

While the H-Representation explanations are minimally complete, their complexity scales with both the number of neurons in the network and the dimensionality of the input space. The latter is particularly unfortunate given that increasing the input dimensions increases the dimensions of each constraint by the same amount.

This means that high-dimensional inputs are less interpretable, given that each
constraint is expressed in terms of all input variables in the form $w1x1 + w2x2...wnxn$.

This limitation could be overcome by introducing \textit{sparsity} to the network, that is to say increasing the number of zeroes in the weight matrix $w$ of each layer. Constraints of this form are more interpretable given that they show more meaningful relationships between individual input variables, rather than all inputs.

\section{Conclusions and Future Work}
In this paper we demonstrated the feasibility of generating exact causal explanations for ReLU neural networks with linear layers, and identified two alternative representations in terms of the H and V Representations. While these explanations are minimally complete with respect to the network's geometry, they are arguably not very satisfying due to the complexity and number of the constraints. The V-Representation offers an alternative which has a constant size with respect to the size of the input, but generating these explanations from the H-Representation introduces exponential complexity with respect to the number of input dimensions. While generating ``Why" explanations is computationally feasible even in the worst case, ``Why not" explanations have exponential complexity in the worst case which may make generating explanations infeasible on certain inputs. This could be overcome by developing polytope marching algorithms which are capable of reducing the search space, such as by identifying super-regions of many polytopes which are known not to contain $N(x)'$. We have discussed explanations from a purely theoretical basis, while the loveliness of explanations is at least partially subjective. In particular, there is a need to gain feedback from real users and regulators, such as in the maritime domain, so as to assess the appropriateness of the explanations to this, or any other domain. We have identified the sparsity of the network and the dimensionality of the input space as the two primary constraints to the ``loveliness" of explanations. The method in this paper could be combined with dimensionality reduction techniques such as Principal Component Analysis (PCA), or techniques that introduce sparsity to the model's weights, such as network pruning, L1 Regression or dropout. While we have demonstrated how to generate explanations for networks consisting of only linear layers, there is a need to expand this to other network architectures such as Convolutional Neural Networks (CNNs). This is a limitation shared with other work on neural network geometry. Depending on the architecture, geometric explanations may in fact be more feasible, given that many introduce sparsity as a design choice. There is also the potential to expand this method to other piecewise linear activation functions, such as Leaky ReLU.

\section*{Acknowledgment}
This work was supported by the UK Hydrographic Office, and the Centre for Assuring Autonomy, a partnership between Lloyd’s Register Foundation and the University of York. We would also like to thank the Maritime and Coastguard Agency for their help and support.

\end{document}